\documentclass{article}
\usepackage{graphicx}
\usepackage{amsmath}
\usepackage{hyperref}
\usepackage{cite}
\usepackage{caption}
\usepackage{tabularx}
\usepackage{array}
\newcolumntype{Z}{>{\raggedright\arraybackslash}X} 
\newcolumntype{Y}{>{\ttfamily\small\raggedright\arraybackslash}X}

\title{Design and Evaluation of Cost-Aware PoQ for Decentralized LLM Inference}

\author{
  Arther Tian\textsuperscript{a},
  Alex Ding\textsuperscript{a,*},
  Frank Chen\textsuperscript{a}\\
  Alan Wu\textsuperscript{a},
  Aaron Chan\textsuperscript{a},
  Bruce Zhang\textsuperscript{a}\\[0.5em]
  \textsuperscript{a}DGrid AI\\[0.5em]
  \textsuperscript{*}Corresponding author: \texttt{alex.ding@dgrid.ai}
}

\date{}

\begin{document}
\maketitle

\begin{abstract}
\noindent
Decentralized large language model (LLM) inference promises transparent and censorship resistant access to advanced AI, yet existing verification approaches struggle to scale to modern models. Proof of Quality (PoQ) replaces cryptographic verification of computation with consensus over output quality, but the original formulation ignores heterogeneous computational costs across inference and evaluator nodes. This paper introduces a cost-aware PoQ framework that integrates explicit efficiency measurements into the reward mechanism for both types of nodes. The design combines ground truth token level F1, lightweight learned evaluators, and GPT based judgments within a unified evaluation pipeline, and adopts a linear reward function that balances normalized quality and cost.

Experiments on extractive question answering and abstractive summarization use five instruction tuned LLMs ranging from TinyLlama-1.1B to Llama-3.2-3B and three evaluation models spanning cross encoder and bi encoder architectures. Results show that a semantic textual similarity bi encoder achieves much higher correlation with both ground truth and GPT scores than cross encoders, indicating that evaluator architecture is a critical design choice for PoQ. Quality–cost analysis further reveals that the largest models in the pool are also the most efficient in terms of quality per unit latency. Monte Carlo simulations over 5\,000 PoQ rounds demonstrate that the cost-aware reward scheme consistently assigns higher average rewards to high quality low cost inference models and to efficient evaluators, while penalizing slow low quality nodes. These findings suggest that cost-aware PoQ provides a practical foundation for economically sustainable decentralized LLM inference.

\end{abstract}

\section{Introduction}

The rapid advancement of large language models (LLMs) has revolutionized artificial intelligence applications, with models such as GPT-4 \cite{openai2023gpt4}, Llama 3 \cite{touvron2023llama}, and Mixtral \cite{jiang2024mixtral} demonstrating unprecedented capabilities in natural language understanding and generation. However, deploying these computationally intensive models in decentralized environments presents significant challenges that traditional centralized architectures do not face \cite{salah2019blockchain}. The convergence of blockchain technology and AI inference promises to democratize access to advanced AI capabilities while ensuring transparency, security, and resistance to single points of failure \cite{zhang2018fhirchain}.

Trustless execution of AI model inference on blockchain networks requires mechanisms to verify both the integrity and quality of computational outputs without relying on trusted third parties. Existing cryptographic approaches such as Zero-Knowledge Machine Learning (ZKML) \cite{feng2021zk} and Optimistic Machine Learning (OPML) \cite{kang2022opml} focus on proving the correctness of inference procedures through circuit-based verification. However, these approaches face severe scalability limitations when applied to modern LLMs containing billions of parameters. For instance, ZKML implementations can only handle models with a few layers, while OPML requires hours to validate even small-scale Transformer models, rendering them impractical for real-world deployment.

\begin{figure}[t]
\centering
\includegraphics[width=\columnwidth]{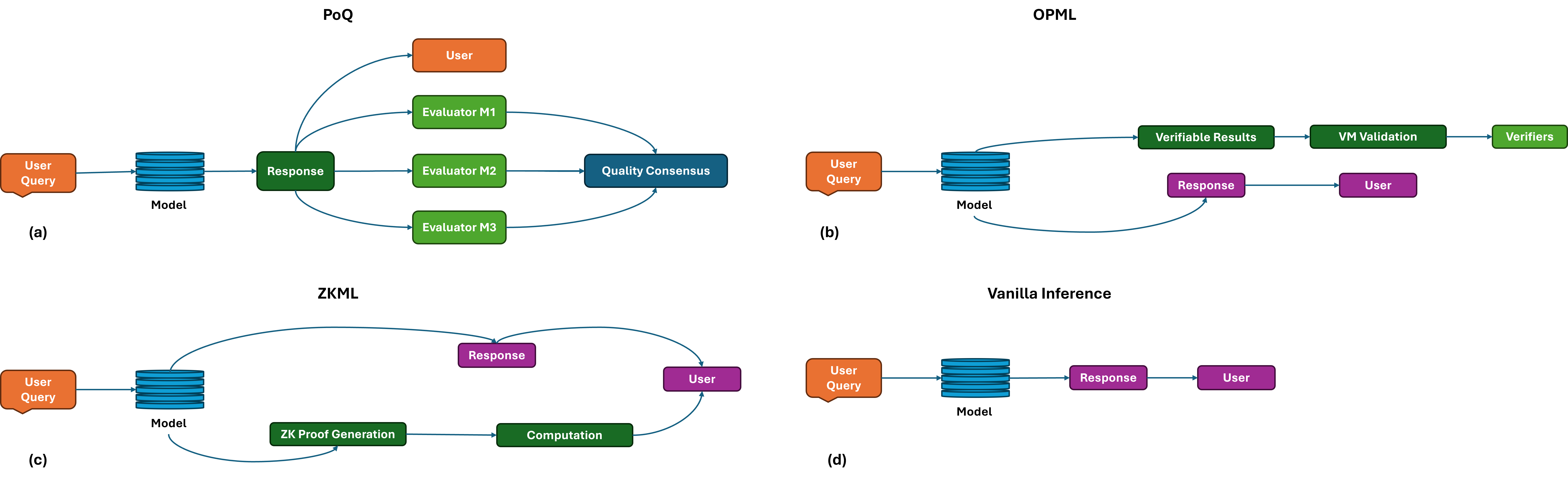}
\caption{Comparison of inference verification paradigms in blockchain environments. (a) Proof of Quality (PoQ) employs multiple lightweight evaluators to assess output quality with minimal overhead. (b) OPML requires expensive VM validation taking minutes to hours. (c) ZKML demands intensive computation for proof generation, often requiring hours for completion. (d) Vanilla inference lacks any verification mechanism, making it unsuitable for trustless environments.}
\label{fig:paradigm_comparison}
\end{figure}

Recently, Zhang et al. proposed Proof of Quality (PoQ) \cite{zhang2024poq}, a novel paradigm that shifts focus from verifying computational processes to assessing output quality. As illustrated in Figure \ref{fig:paradigm_comparison}, PoQ fundamentally differs from existing approaches by employing multiple lightweight evaluation models to assess inference outputs, achieving consensus in seconds rather than hours. This approach leverages cross-encoder models that require orders of magnitude less computation than the original inference, making it suitable for blockchain deployment while maintaining trustworthiness through collective assessment.

Despite its elegance, the original PoQ framework overlooks a critical aspect of decentralized systems: the heterogeneous computational costs across different nodes and models. In practical decentralized networks, inference nodes operate with varying hardware capabilities, energy costs, and model architectures \cite{li2022federated}. Without considering these cost disparities, the incentive mechanism may inadvertently favor computationally expensive models regardless of their quality-to-cost ratio, leading to inefficient resource allocation across the network.

\begin{figure}[htbp]
\centering
\includegraphics[width=\columnwidth]{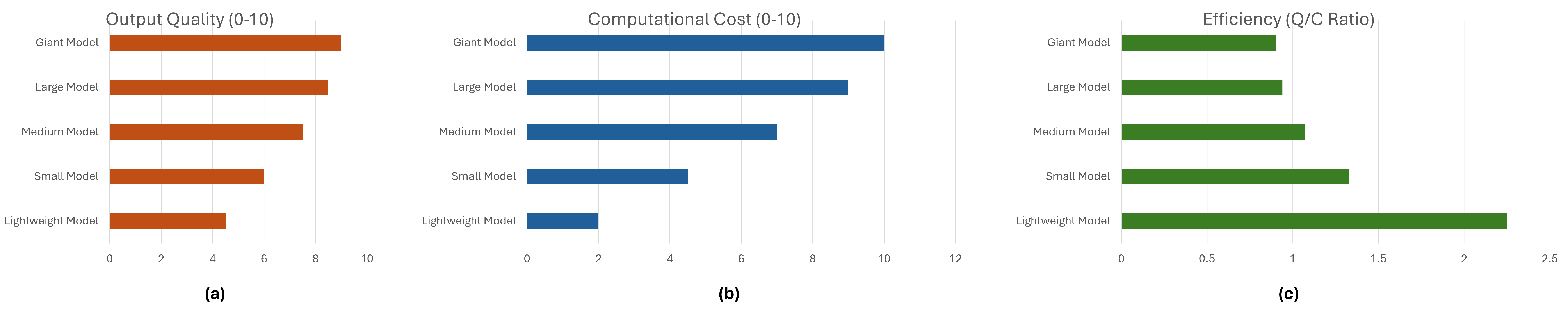}
\caption{Conceptual illustration of the quality-cost trade-off in LLM inference. (a) Output quality increases with model size but exhibits diminishing returns. (b) Computational cost grows disproportionately with model scale. (c) The quality-to-cost efficiency ratio decreases as models become larger, highlighting the need for cost-aware incentive mechanisms. Traditional PoQ rewards based solely on quality would favor giant models despite their poor efficiency.}
\label{fig:quality_cost_tradeoff}
\end{figure}

The relationship between model quality and computational cost is far from linear, as demonstrated in Figure \ref{fig:quality_cost_tradeoff}. While larger models generally achieve higher output quality, their computational costs increase disproportionately, resulting in rapidly diminishing quality-to-cost efficiency. Lightweight models achieve efficiency ratios exceeding 2.0, while giant models drop below 1.0, indicating that doubling computational resources fails to yield proportional quality improvements. This non-linear relationship necessitates a fundamental rethinking of reward mechanisms in decentralized inference networks.

Furthermore, the selection of quality evaluation models significantly impacts both the accuracy of quality assessment and the efficiency of consensus generation. While the original PoQ work primarily focuses on a single cross-encoder model, the choice among multiple evaluator architectures introduces trade-offs between assessment accuracy, computational efficiency, and consensus robustness \cite{reimers2019sentence}. Different evaluator models, i.e., cross-encoders versus bi-encoders, exhibit distinct performance characteristics that must be considered when designing practical PoQ systems.

This paper presents a comprehensive design and evaluation of a cost-aware PoQ mechanism that addresses these fundamental limitations. The proposed framework extends the original PoQ paradigm by incorporating explicit cost considerations into the reward distribution mechanism, ensuring that both inference nodes, denoted as F nodes, and evaluator nodes, denoted as M nodes, are incentivized based on their quality-to-cost efficiency rather than absolute quality alone. The reward function follows the form $R = \alpha \cdot quality - \beta \cdot cost$, where $\alpha$ and $\beta$ are tunable parameters that balance quality and efficiency objectives.

Through extensive experiments on question answering and summarization tasks using five diverse inference models ranging from TinyLlama-1.1B to Llama-3.2-3B, this work demonstrates that cost-aware incentive design leads to more efficient resource utilization while maintaining output quality standards. The evaluation encompasses both objective metrics based on token-level F1 scores and subjective assessments using GPT-4 as a proxy for human judgment, providing comprehensive validation of the proposed approach.

The main contributions of this paper are:

\begin{itemize}
\item A cost-aware extension of the PoQ framework that incorporates computational costs into reward calculations for both inference and evaluator nodes, with parameterized trade-off coefficients enabling flexible optimization for different deployment scenarios.

\item Comprehensive empirical evaluation of multiple lightweight evaluation models including MS-MARCO MiniLM, NLI DeBERTa, and STSB DistilRoBERTa, analyzing their correlation with both ground truth metrics and GPT-based subjective assessments across question answering and summarization tasks.

\item Detailed efficiency profiling of five inference models and three evaluation architectures, quantifying latency, throughput, and GPU memory consumption to establish realistic cost models for decentralized deployments.

\item Monte Carlo simulation of the proposed cost-aware PoQ mechanism over 5,000 rounds, demonstrating improved incentive alignment where efficient models receive proportionally higher rewards despite lower absolute quality scores.

\item Design insights and parameter tuning guidelines for real-world PoQ deployments, including strategies for evaluator selection, consensus threshold setting, and adversarial resilience.
\end{itemize}

\section{Related Work}

This section reviews prior work in three key areas: decentralized AI inference with emphasis on Proof of Quality mechanisms, blockchain-based trustless computing approaches, and quality evaluation methods for generative models.

\subsection{Decentralized AI Inference and PoQ}

The integration of AI inference with blockchain technology has gained significant attention as a means to democratize access to computational resources while ensuring trustworthiness. Salah et al. \cite{salah2019blockchain} provide a comprehensive survey of blockchain applications in AI, identifying key challenges including scalability, energy consumption, and verification overhead. The fundamental challenge lies in ensuring computational integrity without trusted intermediaries, a problem that becomes particularly acute for large-scale models.

Zhang et al. \cite{zhang2024poq} introduced the Proof of Quality paradigm as an alternative to traditional cryptographic verification methods. Unlike approaches that verify computational procedures, PoQ focuses on output quality assessment using lightweight evaluator models. Their framework demonstrates that cross-encoder models can effectively assess LLM outputs with minimal computational overhead, achieving consensus in seconds rather than hours. However, their approach assumes homogeneous costs across nodes and does not account for the heterogeneous nature of decentralized networks.

Distributed AI inference has been explored in federated learning contexts, where Li et al. \cite{li2022federated} analyze challenges in heterogeneous networks including non-IID data distribution, systems heterogeneity, and statistical heterogeneity. While their focus is on training rather than inference, many insights regarding resource allocation and incentive design apply to decentralized inference networks. Similarly, Kang et al. \cite{kang2019incentive} propose incentive mechanisms for federated learning that consider both data quality and computational contributions, though their approach does not extend to inference tasks.

The economic aspects of decentralized AI have been studied by Xiong et al. \cite{xiong2020cloud}, who analyze pricing mechanisms for edge AI services. Their work highlights the importance of balancing quality and cost in resource allocation, though they do not address the specific challenges of trustless environments. Building on game-theoretic foundations, Feng et al. \cite{feng2019joint} design auction mechanisms for distributed machine learning that account for both computational resources and model quality, providing theoretical frameworks that inform cost-aware reward design.

\subsection{Blockchain-Based Trustless Computing}

Cryptographic approaches to verifiable computation have evolved significantly to address the challenge of trustless AI inference. Zero-Knowledge Machine Learning (ZKML) emerged as a promising direction, with Feng et al. \cite{feng2021zk} proposing ZEN, a framework for generating zero-knowledge proofs for neural network inference. However, their approach is limited to small networks with fixed-point arithmetic, making it unsuitable for modern LLMs. Lee et al. \cite{lee2020zknn} extend this work to convolutional neural networks but face similar scalability constraints, requiring hours to generate proofs for models with merely thousands of parameters.

Optimistic Machine Learning (OPML) takes a different approach by assuming honest behavior and only verifying when disputes arise. Kang et al. \cite{kang2022opml} introduce zkCNN for verifiable CNN predictions, achieving better performance than pure ZKML approaches. However, their validation process still requires significant computational resources and time, making real-time inference impractical. The optimistic paradigm has been further explored by Arun et al. \cite{arun2024optimistic}, who propose challenge-based verification for transformer models, though validation latency remains prohibitive for interactive applications.

Alternative verification approaches include Trusted Execution Environments (TEEs), as surveyed by Mofrad et al. \cite{mofrad2021tee}. While TEEs offer hardware-based security guarantees with minimal overhead, they require specialized hardware and trust in manufacturers, limiting their applicability in fully decentralized settings. Tramer and Boneh \cite{tramer2019slalom} combine TEEs with cryptographic techniques to verify neural network inference, achieving better performance than pure cryptographic methods but still requiring trusted hardware.

The consensus mechanisms in blockchain systems directly impact the feasibility of decentralized AI. Xiao et al. \cite{xiao2020survey} survey consensus protocols for blockchain systems, analyzing their trade-offs between security, scalability, and energy efficiency. For AI applications, the consensus overhead must be minimized while maintaining security guarantees, motivating lightweight approaches like PoQ that shift verification from computation to quality assessment.

\subsection{Quality Evaluation for LLM Outputs}

Evaluating the quality of LLM outputs is fundamental to the PoQ paradigm. Traditional metrics such as BLEU \cite{papineni2002bleu} and ROUGE \cite{lin2004rouge} have been widely used but show poor correlation with human judgments for generative tasks. More recent work focuses on learned metrics that better capture semantic similarity and factual accuracy.

Cross-encoder models have emerged as powerful tools for assessing text quality. Reimers and Gurevych \cite{reimers2019sentence} introduce Sentence-BERT, demonstrating that transformer-based models can efficiently compute semantic similarity. Their bi-encoder architecture enables fast similarity computation, though cross-encoders generally achieve higher accuracy. Building on this, Thakur et al. \cite{thakur2021beir} benchmark various retrieval models including cross-encoders on diverse tasks, showing their effectiveness in assessing relevance and quality.

For specific NLP tasks, specialized evaluation methods have been developed. BERTScore \cite{zhang2019bertscore} uses contextual embeddings to evaluate text generation, showing better correlation with human judgments than n-gram based metrics. Similarly, BLEURT \cite{sellam2020bleurt} learns to predict human ratings using synthetic training data, achieving state-of-the-art performance on multiple benchmarks. These learned metrics provide the foundation for quality assessment in PoQ systems.

The use of LLMs as evaluators has gained traction recently. Zheng et al. \cite{zheng2023judging} demonstrate that GPT-4 can serve as a reliable judge for assessing model outputs, showing high agreement with human evaluators. However, they also identify biases such as position bias and verbosity bias that must be addressed. Liu et al. \cite{liu2023gpteval} further analyze the reliability of GPT-based evaluation across different tasks, finding that performance varies significantly with prompt design and task complexity.

Efficiency considerations for evaluation models are crucial for practical deployment. Humeau et al. \cite{humeau2020polyencoders} propose poly-encoders that balance the efficiency of bi-encoders with the accuracy of cross-encoders, achieving 10x speedup with minimal accuracy loss. Similarly, Khattab and Zaharia \cite{khattab2020colbert} introduce ColBERT, which pre-computes document representations while maintaining interaction-based scoring, enabling efficient yet accurate quality assessment.

\section{System Design}

\begin{figure}[htbp]
\centering
\includegraphics[width=0.8\columnwidth]{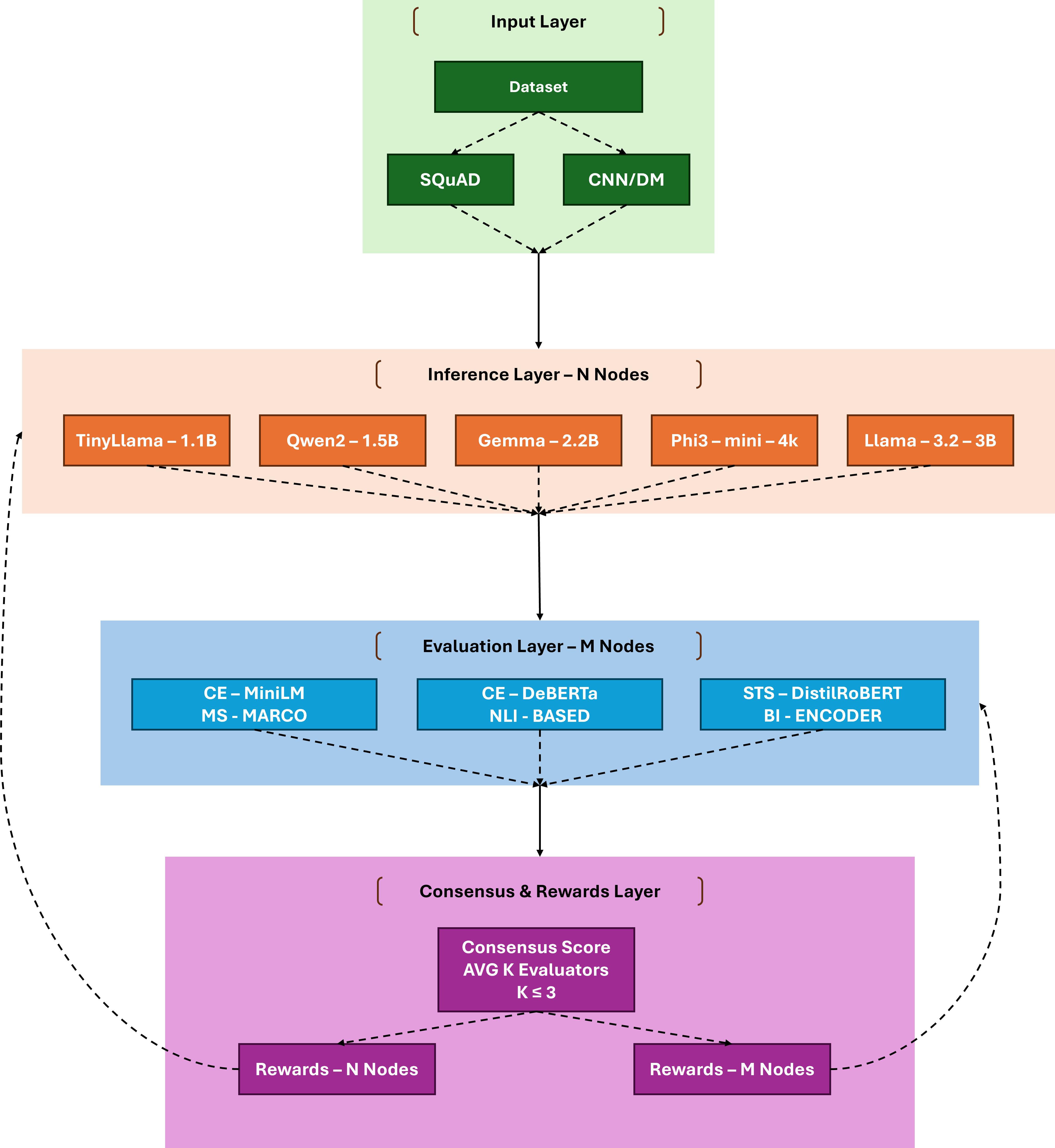}
\caption{Overall architecture of the cost-aware PoQ pipeline. Inputs from SQuAD and CNN/DailyMail are processed by a set of inference nodes denoted as F nodes, evaluated by lightweight models denoted as M nodes, and aggregated in a consensus and rewards layer that computes quality scores and cost-aware incentives.}
\label{fig:system_architecture}
\end{figure}

This section formalizes the system model and presents the cost-aware Proof of Quality (PoQ) framework used to evaluate decentralized large language model inference. The design is organized around four layers: input, inference, evaluation, and consensus with rewards, as illustrated in Figure~\ref{fig:system_architecture}.

\subsection{Problem Setting and Notation}

Let $\mathcal{D} = \{(x_i, y_i, t_i)\}_{i=1}^{N}$ denote the evaluation corpus, where $x_i$ is the input prompt, $y_i$ is the reference answer, and $t_i$ indicates the task type, i.e., question answering or summarization. The corpus is constructed from SQuAD and CNN/DailyMail so that all records share a unified schema.

The system maintains a set of inference nodes
\[
\mathcal{F} = \{f_1, \ldots, f_{|\mathcal{F}|}\},
\]
each node corresponding to a specific large language model. For a sample $i$ and inference node $f \in \mathcal{F}$, the generated output is denoted by $r_{f,i}$.

Ground-truth style quality is measured by a token-level F1 metric $G(x_i, y_i, r_{f,i})$ mapped to the range $[0, 10]$. This value is pre-computed offline for every pair $(f, i)$ and later used to analyze correlation and incentive alignment but not to drive on-chain rewards.

A second set of nodes contains lightweight evaluators
\[
\mathcal{M} = \{m_1, \ldots, m_{|\mathcal{M}|}\},
\]
each node corresponding to a small encoder or cross-encoder model. For a record $(x_i, r_{f,i})$ and evaluator $m$, the raw similarity score is denoted by $s_{m,f,i}$. Since different evaluators produce scores on incomparable scales, all raw scores are transformed into normalized values
\[
e_{m,f,i} \in [0, 10]
\]
through min–max normalization over the full evaluation set. These normalized evaluator scores serve as the basic quality signals inside PoQ.

To model heterogeneous hardware and latency, each inference node $f$ is associated with a cost value $C_F(f) \in [0,1]$ and each evaluator node $m$ is associated with $C_M(m) \in [0,1]$. Costs are derived from measured average latency and normalized across nodes. Additional statistics such as throughput and peak memory are recorded for analysis in later sections but only latency-based cost is used inside the reward function.

\subsection{Cost-Aware PoQ Framework}

The cost-aware PoQ framework extends the original PoQ paradigm~\cite{zhang2024poq} by explicitly incorporating node costs into the reward calculation. For each PoQ round, the protocol selects one data record and one inference node, then aggregates the opinions of a small subset of evaluators.

Given a sample index $i$ and an inference node $f$, a subset of evaluators
\[
S_{i,f} \subseteq \mathcal{M}
\]
is chosen such that $1 \leq |S_{i,f}| = K \leq 3$. All selected evaluators have pre-computed normalized scores $e_{m,f,i}$ for the pair $(x_i, r_{f,i})$.

The consensus quality score for this round is defined as
\begin{equation}
Q_{i,f} = \frac{1}{10K} \sum_{m \in S_{i,f}} e_{m,f,i},
\label{eq:consensus_quality}
\end{equation}
which lies in $[0,1]$ by construction. This value captures the average opinion of the sampled evaluators regarding the output quality of node $f$ on record $i$.

The reward for the inference node combines the consensus quality with its normalized cost:
\begin{equation}
R_F(i,f) = \alpha_F \, Q_{i,f} - \beta_F \, C_F(f),
\label{eq:reward_f}
\end{equation}
where $\alpha_F > 0$ and $\beta_F > 0$ control the trade-off between quality and efficiency. When $\beta_F$ increases, high-latency models are penalized more strongly, so that efficient models can receive larger rewards even with slightly lower absolute quality.

Evaluator nodes are rewarded based on how closely their scores agree with the consensus while also accounting for their own costs. For each $m \in S_{i,f}$, define a deviation term
\begin{equation}
d_{i,f,m} = \frac{1}{10} \left| e_{m,f,i} - \frac{1}{K} \sum_{m' \in S_{i,f}} e_{m',f,i} \right|,
\end{equation}
which lies in $[0,1]$ and equals zero when the evaluator perfectly matches the average of its peers. The corresponding closeness score is
\begin{equation}
C^{\text{close}}_{i,f,m} = 1 - d_{i,f,m}.
\end{equation}
The reward for evaluator $m$ is then
\begin{equation}
R_M(i,f,m) = \alpha_M \, C^{\text{close}}_{i,f,m} - \beta_M \, C_M(m),
\label{eq:reward_m}
\end{equation}
with $\alpha_M > 0$ and $\beta_M > 0$ analogous to the coefficients for inference nodes. Evaluators that consistently disagree with the consensus incur large deviations, which reduces their rewards even if their computational cost is low. This design encourages evaluators that are both accurate and efficient.

\subsection{Inference and Evaluator Nodes}

The inference layer contains five heterogeneous F nodes, each representing a publicly available large language model with different parameter counts and architectural choices. The models span from compact variants such as TinyLlama with $1.1$ billion parameters to larger models such as Llama 3.2 with $3$ billion parameters. All models are prompted using a unified instruction template tailored to the task type, i.e., extractive question answering or abstractive summarization. For each record $x_i$ and each inference node $f$, the generated output $r_{f,i}$ is produced once offline and stored in a generation log that is reused during PoQ simulation.

The evaluation layer contains three lightweight M nodes. Two nodes are cross-encoder models based on MiniLM trained on MS MARCO style relevance labels and DeBERTa trained for natural language inference. The third node is a bi-encoder model based on DistilRoBERTa trained on semantic textual similarity data. Cross-encoders compute fine-grained token interactions and typically achieve higher accuracy, while the bi-encoder design offers improved efficiency since representations can be pre-computed. All three evaluators operate on the pair $(x_i, r_{f,i})$ and produce raw similarity scores that are subsequently normalized to $[0,10]$ as described earlier.

To instantiate the cost model, each inference and evaluation node is profiled on a common GPU. The profiler measures average latency and throughput over randomly sampled records. Latency values are linearly scaled into $[0,1]$ to obtain $C_F(f)$ and $C_M(m)$. These costs are then treated as fixed parameters inside the PoQ framework so that the simulation can focus on the impact of the reward function rather than low-level hardware variability.

\subsection{Consensus Mechanism}

The consensus and rewards layer implements the cost-aware PoQ protocol on top of the pre-computed generation and evaluation logs. A single PoQ round proceeds according to the following steps.

\textbf{Algorithm 1: Single round of cost-aware PoQ.}
\begin{enumerate}
    \item Sample a record index $i$ uniformly from $\{1, \ldots, N\}$.
    \item Select an inference node $f$ according to a scheduling policy; in the simulation this choice is uniform over $\mathcal{F}$.
    \item Retrieve the cached output $r_{f,i}$ and the corresponding evaluator scores $e_{m,f,i}$ for all $m \in \mathcal{M}$ that produced a score on this record.
    \item Sample a subset $S_{i,f}$ of evaluators without replacement such that $1 \leq |S_{i,f}| = K \leq 3$.
    \item Compute the consensus quality $Q_{i,f}$ from Equation~\eqref{eq:consensus_quality}.
    \item For the selected inference node, compute the reward $R_F(i,f)$ using Equation~\eqref{eq:reward_f} and add it to the cumulative reward statistics of node $f$.
    \item For each evaluator $m \in S_{i,f}$, compute the deviation $d_{i,f,m}$, the closeness score $C^{\text{close}}_{i,f,m}$, and the reward $R_M(i,f,m)$ from Equation~\eqref{eq:reward_m}, then update the cumulative statistics of node $m$.
\end{enumerate}

Running this procedure for many rounds produces empirical reward distributions for both inference and evaluation nodes. Since all heavy computation is performed offline, the simulation can explore different values of $\alpha_F$, $\beta_F$, $\alpha_M$, $\beta_M$, and $K$ without repeating LLM inference or evaluator scoring. Subsequent sections analyze how these design choices influence quality, efficiency, and incentive alignment in decentralized PoQ deployments.

\section{Experimental Setup}

This section describes the datasets, models, evaluation protocol, and efficiency profiling procedure used to study cost-aware Proof of Quality (PoQ) in decentralized LLM inference.

\subsection{Datasets and Models}

\subsubsection*{Tasks and datasets}

Two representative natural language generation tasks are considered: extractive question answering and abstractive summarization. For question answering, the development split of SQuAD v1.1 is used, which contains 10\,570 human annotated questions on Wikipedia articles \cite{rajpurkar-etal-2016-squad}. For summarization, the anonymized CNN/DailyMail news corpus is employed, which provides article–highlights pairs derived from large scale news stories \cite{hermann2015teaching}.

Both datasets are converted into a unified task format. Each example contains a unique identifier, an input field, and a reference field. For SQuAD, the input concatenates the question and the supporting context paragraph, while the reference holds the ground truth answer span. For CNN/DailyMail, the input corresponds to the news article and the reference contains the human written highlights. From each dataset, 200 examples are sampled using the data construction scripts so that the total evaluation corpus consists of 400 prompts spanning both tasks.

\begin{table}[t]
\centering
\small
\caption{Datasets and tasks used in the experiments. The sample counts correspond to the constructed evaluation corpus.}
\label{tab:datasets}
\begin{tabularx}{\columnwidth}{l X c c c X}
\hline
Dataset & Task type & Split & Original size & Samples used & Ground truth metric \\
\hline
SQuAD v1.1 & Extractive question answering & Dev  & 10\,570 & 200 & Token level F1 scaled to $[0,10]$ \\
CNN/DailyMail & Abstractive summarization & Test & 11\,490 & 200 & Token level F1 scaled to $[0,10]$ \\
\hline
\end{tabularx}
\end{table}

\subsubsection*{Inference models (F nodes)}

Five open source instruction tuned LLMs serve as inference nodes, matching the design in Figure~\ref{fig:system_architecture}. The models are loaded via the Transformers library and are treated as black box generators whose internal parameters are not modified. Their configuration is encoded in a model registry shared by the generation and efficiency scripts.

\begin{table}[t]
\centering
\small
\caption{Inference models used as F nodes. Parameter scales follow the respective model cards and indicate the order of magnitude.}
\label{tab:inference_models}
\begin{tabular}{l l c}
\hline
Node key & Model name & Scale \\
\hline
tinyllama\_1\_1b & TinyLlama-1.1B-Chat-v1.0 & 1.1B \\
qwen2\_1\_5b     & Qwen2-1.5B-Instruct       & 1.5B \\
gemma\_2\_2b\_it & Gemma-2-2B-it             & 2.2B \\
phi3\_mini\_4k   & Phi-3-mini-4k-instruct    & 3.8B \\
llama\_3\_2\_3b  & Llama-3.2-3B-Instruct     & 3.0B \\
\hline
\end{tabular}
\end{table}

All inference models are executed using the text generation pipeline with identical decoding configuration, including a fixed maximum of 128 new tokens, nucleus sampling with moderate temperature, and deterministic random seeds, as specified in the answer generation script. The efficiency script measures per sample latency with batch size equal to one for all generation models so that latency and throughput reflect single request performance.

\subsubsection*{Evaluation models (M nodes)}

The PoQ framework relies on lightweight evaluation models that estimate semantic quality of candidate outputs. Three such models are instantiated as evaluator nodes.

\begin{itemize}
  \item \textbf{CE--MiniLM}: a cross encoder based on a MiniLM model trained on the MS MARCO passage ranking dataset \cite{nguyen2016msmarco}. The model processes concatenated reference–candidate pairs and returns a relevance score.
  \item \textbf{CE--DeBERTa}: a cross encoder based on a DeBERTa v3 small model trained on natural language inference data. The model estimates whether the candidate answer semantically entails the reference answer.
  \item \textbf{STS--DistilRoBERTa}: a bi encoder based Sentence Transformer trained on the STS Benchmark and related semantic similarity corpora \cite{cer-etal-2017-semeval}. It encodes reference and candidate separately and computes cosine similarity.
\end{itemize}

Table~\ref{tab:eval_models} lists the evaluator configuration, including the batch sizes used during efficiency profiling.

\begin{table}[t]
\centering
\small
\caption{Evaluation models used as M nodes and their efficiency profiling batch sizes.}
\label{tab:eval_models}
\begin{tabular}{l l l c}
\hline
Node key & Model name & Architecture & Batch size \\
\hline
ce\_minilm  & MiniLM MS~MARCO     & Cross encoder & 32 \\
ce\_deberta & DeBERTa NLI         & Cross encoder & 32 \\
sts\_stsb   & DistilRoBERTa STS-B & Bi encoder    & 64 \\
\hline
\end{tabular}
\end{table}

\subsection{Evaluation Methodology}

The experimental pipeline is structured into four stages implemented by scripts under the project source tree.

\subsubsection*{Answer generation}

First, a unified task file is constructed by merging the processed SQuAD and CNN/DailyMail samples. The answer generation script iterates over all prompts and models, uses the text generation pipeline to produce a single response for each prompt, and writes one generation log per model. Each record stores the dataset name, task type, example identifier, model key, prompt, reference text, and generated output.

\subsubsection*{Ground truth scoring}

Objective ground truth scores are computed by the metric construction script. For each record, token level precision and recall are calculated between the generated output and the reference text using the standard span based F1 definition from SQuAD:
\[
\mathrm{F1} = \frac{2 \cdot \mathrm{precision} \cdot \mathrm{recall}}{\mathrm{precision} + \mathrm{recall}}.
\]
The raw F1 value lies in $[0,1]$ and is multiplied by ten to obtain a score in $[0,10]$ that is compatible with PoQ rewards. The script writes the resulting ground truth score back into each generation record and aggregates all records into a consolidated metric file.

\subsubsection*{Lightweight evaluator scoring}

Quality estimates from the evaluation models are obtained in two stages. The evaluator scoring script runs the two cross encoder models on every reference–output pair. For each evaluator, raw similarity scores are collected and then normalized to the range $[0,10]$ by per task min–max scaling so that low quality outputs receive scores near zero and high quality outputs approach ten. The normalized values are stored as dedicated fields in the metric file.

To complement these scores with a bi encoder perspective, the correlation analysis script loads the merged records and applies the STS evaluator. Reference and candidate sentences are encoded with the STS model, cosine similarity is computed, and a second min–max transformation maps similarities to $[0,10]$. The resulting fields provide an additional quality signal that is used both for correlation analysis and for extended PoQ reward experiments.

\subsubsection*{GPT based evaluation}

Subjective evaluation is performed by an LLM judge based on the \texttt{gpt-4o-mini} model. The judge set construction script samples up to $N_{\text{group}} = 30$ examples for each combination of dataset, task type, and inference model from the metric file. For every selected record, the script reconstructs the original input and reference text and stores them together with the model output in a dedicated labeling set.

The judging script sends these items to the OpenAI Responses API with a fixed system prompt that instructs the judge to rate each candidate answer from 0 to 10 and to return a short justification in a JSON object of the form
\texttt{\{"score": number, "justification": "..."\}}. The returned scores are parsed, validated, and written to a labeling result file. Finally, a merging script joins GPT scores back into the main metric file and produces summary statistics that include the number of judged samples and the mean and standard deviation of GPT scores per model and task.

\subsection{Efficiency Profiling}

Efficiency measurements quantify the computational cost of both inference and evaluation nodes. All profiling runs are executed on a single workstation equipped with an NVIDIA RTX 4090 GPU, an Intel Core i9-12900K CPU, and 64\,GB of system memory. Experiments use CUDA enabled PyTorch with automatic selection of the \texttt{bfloat16} data type when a GPU is available.

The profiling procedure is implemented in the efficiency measurement script. A random subset of prompts is drawn from the unified task file for each model, with the number of examples per model controlled by a configuration parameter whose default value equals 100. For inference models, the script constructs generation prompts in the same format as the main experiments, warms up the model with three prompts to amortize initialization overhead, and then times a single pass over the sampled prompts using batch size one. After synchronization, total elapsed time and peak GPU memory usage are recorded, and the script reports average per sample latency in milliseconds, throughput in samples per second, and maximum memory footprint in megabytes.

For evaluation models, the script creates reference–candidate pairs from the same subset of prompts and timing is performed separately for each evaluator. Cross encoder models use their configured batch size, i.e., 32 for CE–MiniLM and CE–DeBERTa, while the STS bi encoder uses batch size 64. Each evaluator performs a short warmup phase followed by a full measurement pass with GPU synchronization and memory statistics. The resulting efficiency records are written to a summary file that directly feeds into the cost model used by the PoQ simulation.

This unified setup ensures that all quality and cost measurements are derived from a consistent corpus, model configuration, and hardware environment, enabling a faithful study of the quality–cost trade offs inherent in decentralized PoQ.

\section{Results}

This section describes empirical findings on evaluator reliability, quality–cost trade offs for inference models, and incentive alignment under cost-aware Proof of Quality (PoQ) in decentralized LLM inference. All results are computed on the unified evaluation corpus described in Experimental Setup section.

\subsection{Evaluator Correlation Analysis}

\begin{figure}[t]
    \centering
    \includegraphics[width=\columnwidth]{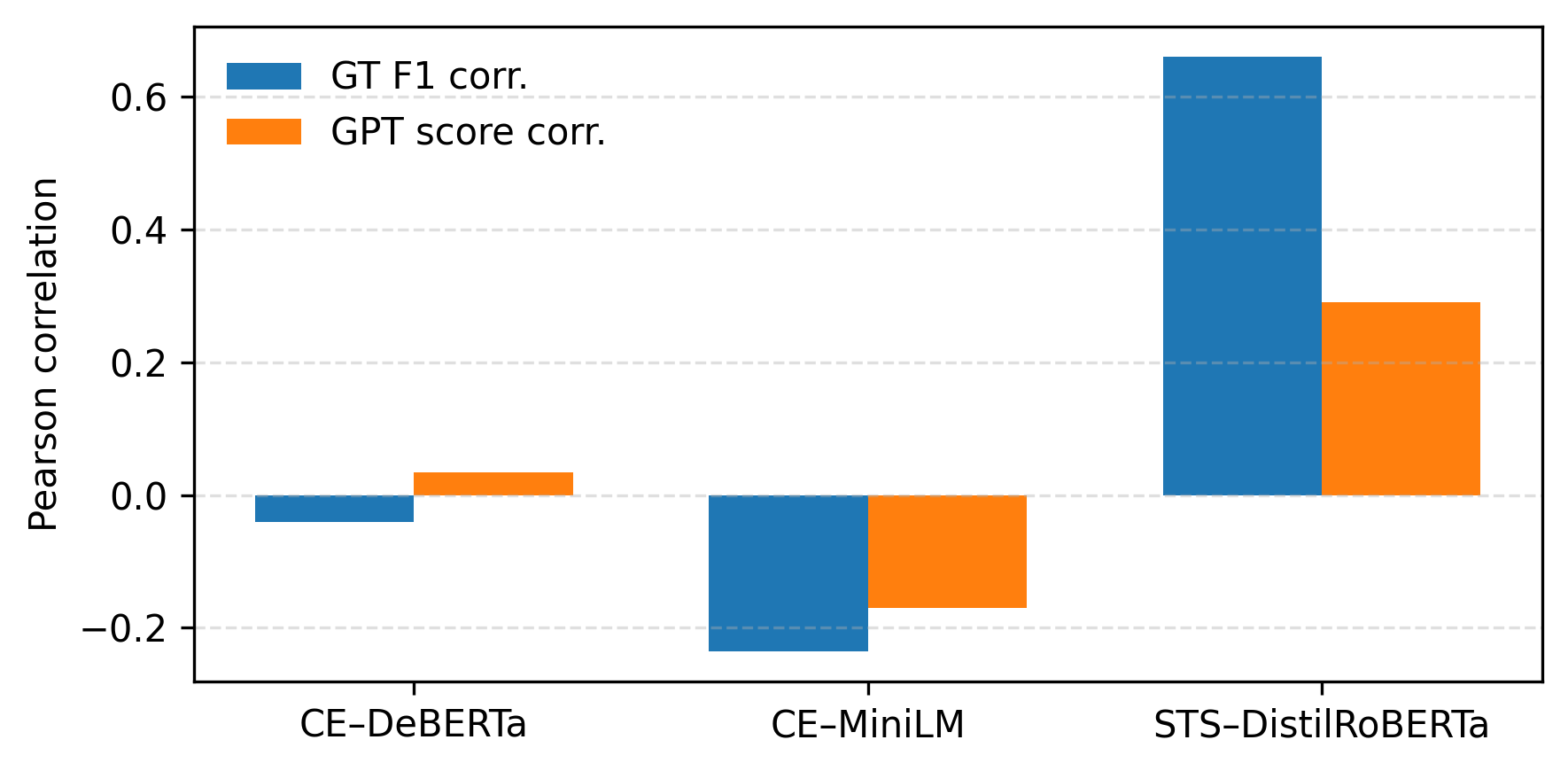}
    \caption{Correlation between evaluator scores and reference signals. Bars show Pearson correlation with ground truth F1 scores and GPT based judgments, averaged over question answering and summarization tasks.}
    \label{fig:eval_correlations}
\end{figure}

Figure~\ref{fig:eval_correlations} summarizes the Pearson correlation between scores from the three lightweight evaluators and two reference signals, i.e., token level F1 against ground truth answers and GPT based judgments. The correlation values are averaged over question answering and summarization tasks.

The STS--DistilRoBERTa bi encoder achieves the strongest alignment with both references. Its scores reach approximately $0.66$ correlation with ground truth F1 and $0.29$ correlation with GPT scores, indicating that semantic similarity in the STS embedding space captures much of the variance in both objective and subjective quality assessments. In contrast, the two cross encoders show correlations close to zero or even negative. CE--DeBERTa obtains roughly $-0.04$ correlation with ground truth F1 and $0.03$ with GPT scores, while CE--MiniLM exhibits moderately negative correlations of about $-0.24$ and $-0.17$ respectively.

These results indicate that not all learned quality metrics are equally suitable as PoQ evaluators. Despite being trained on relevance and textual entailment data, the cross encoders fail to provide stable quality signals for the LLM outputs considered here. The STS model, although architecturally simpler, yields more consistent rankings across tasks and reference types. The correlation analysis therefore supports the use of STS--DistilRoBERTa as a primary evaluator in subsequent experiments, with cross encoders serving mainly as diversity sources in multi evaluator configurations.

\subsection{Quality--Cost Trade Offs of Inference Models}

\begin{figure}[t]
    \centering
    \includegraphics[width=\columnwidth]{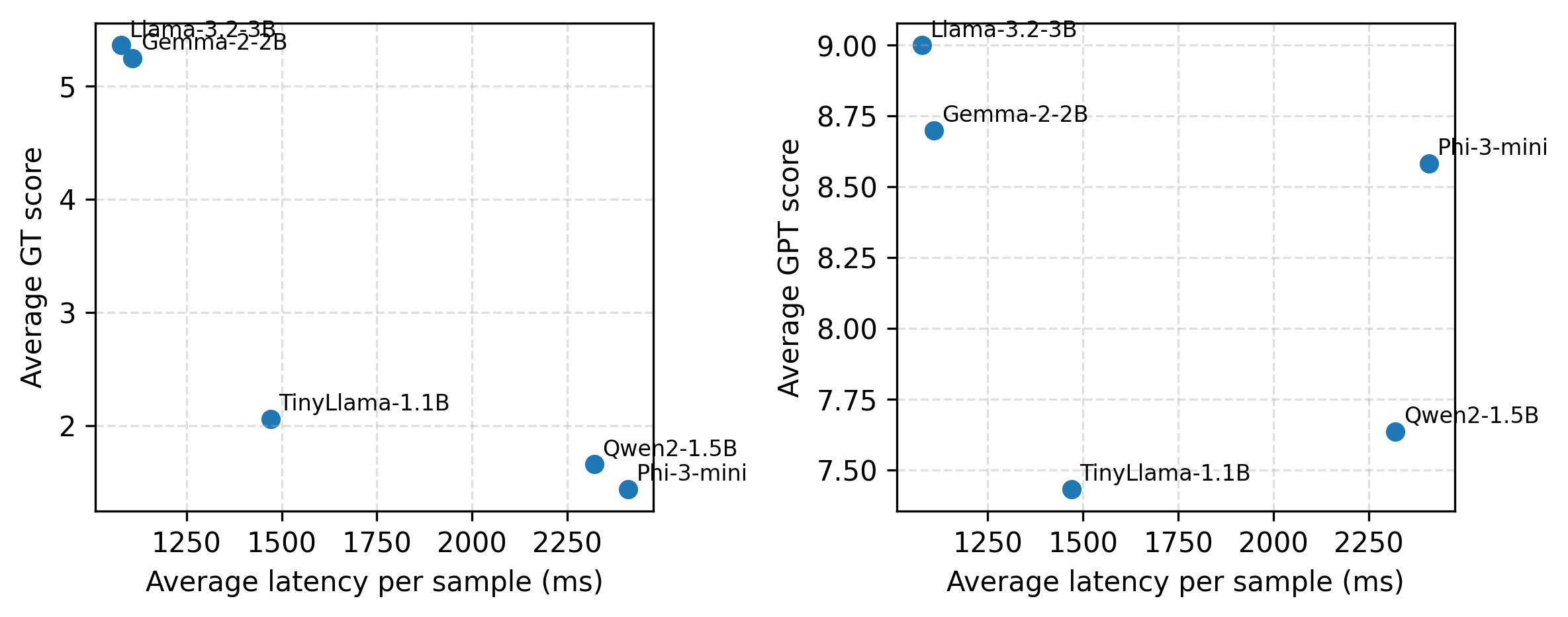}
    \caption{Quality--cost trade offs for inference models. Left: average ground truth F1 score versus latency. Right: average GPT based score versus latency. Each point corresponds to a single inference model.}
    \label{fig:quality_cost_tradeoffs}
\end{figure}

Figure~\ref{fig:quality_cost_tradeoffs} plots average quality against per sample latency for the five inference models. The left panel uses the aggregated ground truth F1 score as quality metric, averaged over the two tasks, while the right panel reports average GPT judgment scores on the subsampled evaluation subset.

Llama-3.2-3B and Gemma-2-2B form the high quality frontier. Both models achieve average ground truth scores slightly above $5.3$ on the $[0,10]$ scale and average GPT scores of approximately $9.0$ and $8.7$ respectively. At the same time their average per sample latency remains near $1.1$ seconds. TinyLlama-1.1B lies in the middle region, delivering an average ground truth score of about $2.1$ and a GPT score around $7.4$ with latency around $1.47$ seconds.

Phi-3-mini and Qwen2-1.5B occupy the low quality and high cost corner. Their average ground truth scores stay below $1.7$ while GPT scores remain between $7.6$ and $8.6$. However, both models require more than $2.3$ seconds per sample, i.e., more than double the latency of the two larger models. When normalizing ground truth quality by latency, Llama-3.2-3B obtains roughly seven times higher quality per millisecond than Qwen2-1.5B and Phi-3-mini, with Gemma-2-2B close behind.

The scatter plots highlight that larger parameter count does not necessarily imply worse efficiency. In this setup, the two largest models achieve both higher absolute quality and better quality--cost efficiency than several smaller competitors. This observation motivates the need for incentive mechanisms that reward quality relative to cost rather than parameter count alone, since naïve assumptions about model size and efficiency can be misleading in heterogeneous hardware environments.

\subsection{PoQ Simulation Outcomes}

\begin{figure}[t]
    \centering
    \includegraphics[width=\columnwidth]{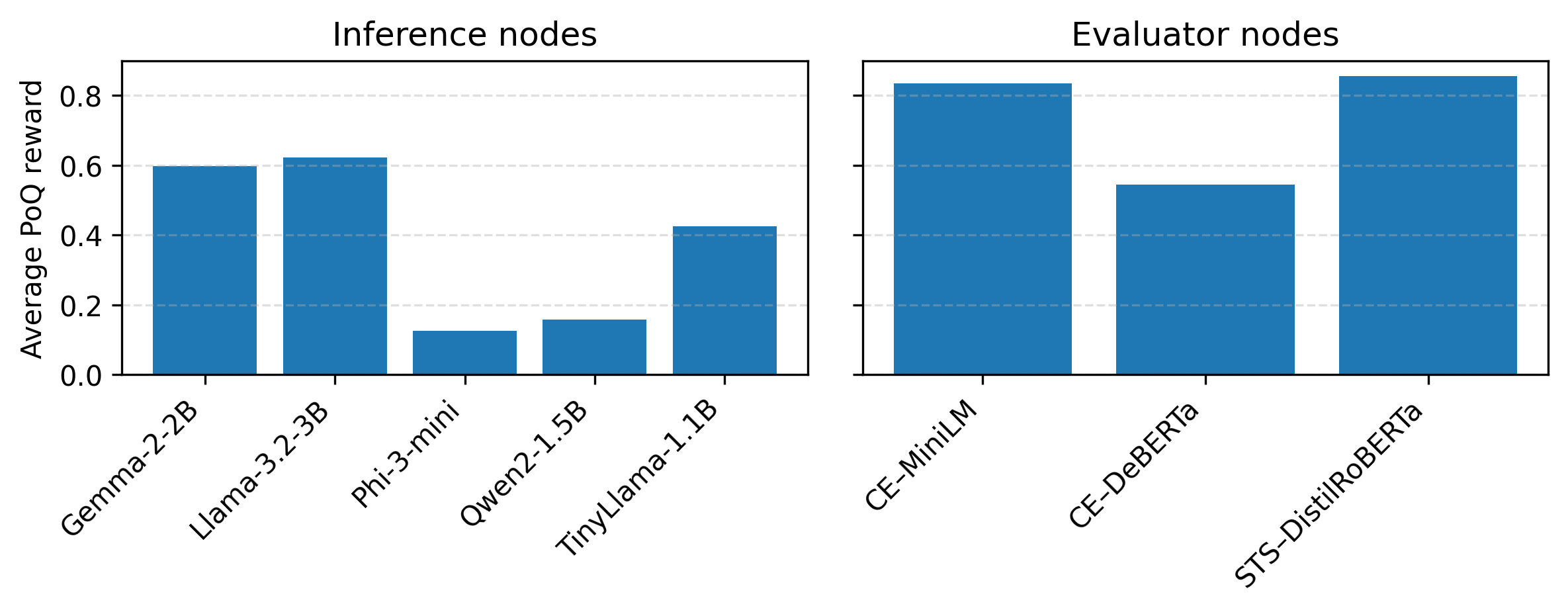}
    \caption{Average PoQ rewards from Monte Carlo simulation. Left: inference nodes. Right: evaluator nodes. Rewards are reported in normalized units.}
    \label{fig:poq_rewards}
\end{figure}

To evaluate incentive alignment under cost-aware PoQ, Monte Carlo simulations are run over 5\,000 consensus rounds using the quality and cost measurements described above. Table~\ref{tab:poq_nodes} reports summary statistics for each node and Figure~\ref{fig:poq_rewards} visualizes the resulting average rewards for inference and evaluator nodes.

\subsubsection*{Inference nodes}

Among inference nodes, Llama-3.2-3B attains the highest average reward, approximately $0.62$ in normalized units, followed closely by Gemma-2-2B at $0.60$. Both nodes combine high ground truth and GPT based quality as indicated in Figure~\ref{fig:quality_cost_tradeoffs}, together with low normalized cost. The cost normalization places Llama-3.2-3B at the cheapest end of the latency spectrum, while Gemma-2-2B incurs only slightly higher latency.

TinyLlama-1.1B earns a moderate average reward of about $0.43$ despite lower absolute quality. Its ground truth and GPT scores occupy the middle region of the quality frontier, yet the model enjoys a significantly lower cost score than Phi-3-mini and Qwen2-1.5B. This behavior illustrates the intended effect of the cost-aware reward design, i.e., models that deliver reasonable quality at reduced cost remain competitive.

Phi-3-mini and Qwen2-1.5B receive the lowest rewards, approximately $0.13$ and $0.16$. Both models exhibit low quality and high normalized cost due to their large latency overhead. Their job counts are comparable to those of other inference nodes, yet the cost-aware PoQ mechanism consistently demotes them in reward ranking. Overall, the inference side of the simulation demonstrates that quality-to-cost efficiency rather than raw quality dominates reward outcomes.

\subsubsection*{Evaluator nodes}

Evaluator nodes show a different but related pattern. STS--DistilRoBERTa and CE--MiniLM achieve the highest average rewards, about $0.86$ and $0.83$, while CE--DeBERTa lags behind at roughly $0.54$. The cost normalization assigns the lowest cost to STS--DistilRoBERTa, followed very closely by CE--MiniLM, since both evaluators process batched inputs with sub-millisecond latency per pair. CE--DeBERTa exhibits significantly higher latency, which translates into a cost norm of one and reduces its net reward despite playing the same evaluation role.

Combined with the correlation analysis, these results suggest that STS--DistilRoBERTa is simultaneously the most informative and the most efficient evaluator, hence the most attractive choice for decentralized deployments. CE--MiniLM, although weakly correlated with ground truth and GPT scores, still receives substantial reward due to very low cost and the additional diversity it brings to the evaluator set. CE--DeBERTa, in contrast, underperforms on both axes and therefore becomes disfavored by the PoQ incentive structure.

\subsubsection*{Summary}

Taken together, the simulation confirms that cost-aware PoQ can steer both inference and evaluation nodes toward desirable quality--cost trade offs. High quality and low latency inference models rise to the top of the reward distribution, efficient evaluators with strong alignment to reference signals are preferred, and high cost low quality models receive consistently lower compensation even when assigned a similar number of jobs. The mechanism therefore offers a practical path toward economically sustainable decentralized LLM inference where heterogeneous nodes are rewarded according to their actual contribution.

\begin{table}[t]
\centering
\small
\caption{Cost-aware PoQ simulation results for inference and evaluator nodes. Average rewards are reported in normalized units. Cost values are normalized to $[0,1]$.}
\label{tab:poq_nodes}
\begin{tabular}{l l c c c r}
\hline
Node type & Node key & Avg reward & Avg latency (ms) & Cost norm & Jobs \\
\hline
inference & gemma\_2\_2b\_it  & 0.598 & 1108.0 & 0.023 & 952 \\
inference & llama\_3\_2\_3b   & 0.623 & 1077.7 & 0.000 & 1009 \\
inference & phi3\_mini\_4k    & 0.126 & 2409.3 & 1.000 & 1037 \\
inference & qwen2\_1\_5b      & 0.157 & 2320.6 & 0.933 & 1039 \\
inference & tinyllama\_1\_1b  & 0.426 & 1470.1 & 0.295 & 963 \\
\hline
eval      & ce\_minilm        & 0.834 &    1.0 & 0.007 & 5000 \\
eval      & ce\_deberta       & 0.544 &    5.9 & 1.000 & 5000 \\
eval      & sts\_stsb         & 0.856 &    0.9 & 0.000 & 5000 \\
\hline
\end{tabular}
\end{table}

\section{Discussion}

The empirical results highlight several design lessons for cost-aware Proof of Quality (PoQ) in decentralized LLM inference, as well as practical limitations of the current study.

\subsection{Implications for PoQ Design}

The first implication concerns the role of evaluator choice. The original PoQ formulation focuses on cross-encoder style quality estimators \cite{zhang2024poq}, motivated by their strong performance on semantic matching benchmarks \cite{reimers2019sentence}. In the present setting, however, the two cross encoders exhibit near-zero or even negative correlation with both ground truth F1 and GPT based judgments, whereas the STS--DistilRoBERTa bi encoder achieves substantially higher alignment. This outcome suggests that evaluator architectures tuned for retrieval or natural language inference do not automatically transfer to LLM generation quality assessment. In practice, bi encoder models trained on semantic textual similarity data appear more robust for PoQ, especially when cost constraints favor compact architectures.

A second implication relates to the interaction between cost-aware rewards and model scaling. The quality--cost frontier in Figure~\ref{fig:quality_cost_tradeoffs} shows that parameter count alone is a poor proxy for efficiency. Llama-3.2-3B and Gemma-2-2B dominate both quality and latency, while smaller models such as Qwen2-1.5B and Phi-3-mini deliver lower quality at significantly higher cost. The PoQ simulation reflects this structure: reward allocation favors the larger but more efficient models, while high-cost low-quality models are consistently penalized. This behavior indicates that cost-aware PoQ can implicitly perform a form of hardware-aware model selection, shifting demand toward architectures that exploit the underlying accelerator more effectively.

The third implication concerns the interaction between inference nodes and evaluator nodes. The reward distribution in Figure~\ref{fig:poq_rewards} shows that efficient evaluators with strong correlation to reference signals, exemplified by STS--DistilRoBERTa, can earn rewards on par with or even exceeding those of high-quality inference nodes. This observation underlines an important aspect of decentralized AI marketplaces: supplying reliable evaluation services can be economically competitive with running large inference models, as long as evaluators remain lightweight and informative. In addition, the fact that CE--MiniLM still receives non-trivial rewards despite weak correlation indicates that the PoQ mechanism values diversity among evaluators. A heterogeneous evaluator set can mitigate correlated biases of any single model, similar in spirit to ensemble methods in classical machine learning.

\subsection{Guidelines for Practical Deployment}

The findings suggest several practical guidelines for deploying cost-aware PoQ in real systems. For evaluator selection, semantic similarity models trained on STS style data should be considered as default candidates, while cross encoders can be added selectively to increase architectural diversity rather than to serve as primary judges. Evaluator latency needs to be measured under realistic batch sizes, since throughput gains from larger batches can substantially reduce effective cost, especially on modern GPUs.

For inference node selection and incentive tuning, the experiments indicate that quality and cost should be normalized within the pool of available models rather than by absolute thresholds. The reward function used in the experiments, where a linear combination of normalized quality and cost determines payouts, can be tuned by adjusting the relative weights assigned to these components. In environments where latency and energy consumption are critical, the cost coefficient can be increased to further penalize slow or resource-intensive models. Conversely, settings that prioritize accuracy can adopt a higher quality coefficient while still maintaining a baseline level of cost sensitivity.

The simulation also highlights the importance of transparent cost measurement. All latency and memory figures are collected on a fixed hardware configuration, which allows clear comparison across models. In heterogeneous networks, participating nodes may run on different accelerators and incur different energy prices. A practical deployment would therefore benefit from on-chain or commit-reveal protocols that standardize cost reporting, possibly combined with random auditing to deter misreporting. Cost-aware PoQ does not remove the need for such mechanisms, but provides a principled way to translate reported costs into economic incentives.

\subsection{Limitations and Future Directions}

Several limitations of the current study point to opportunities for future work. First, the evaluation focuses on English language question answering and summarization tasks with relatively short contexts. Real-world decentralized inference markets are likely to serve multilingual users and long-context applications such as document understanding or multi-turn dialogue. The quality--cost trade offs observed here may change under different task distributions or sequence lengths, especially for models designed explicitly for long-context processing.

Second, the experimental setup employs a single hardware platform with one high-end GPU. While this choice isolates differences between models, it does not capture the full variability of decentralized networks in which nodes may operate on consumer GPUs, CPUs, or specialized accelerators. Extending the cost model to include heterogeneous hardware, energy prices, and network bandwidth costs would provide a more realistic view of incentive alignment. Techniques from federated learning on heterogeneous devices \cite{li2022federated} and pricing models for edge computing services \cite{xiong2020cloud} offer useful starting points for such extensions.

Third, the PoQ simulation assumes honest behavior by inference and evaluator nodes, aside from natural quality variation. Adversarial strategies, i.e., collusion between evaluators and inference nodes or targeted manipulation of evaluator outputs, are not modeled. Prior work on LLM-as-a-judge has documented position bias and other systematic effects in GPT-based evaluation \cite{zheng2023judging,liu2023gpteval}, indicating that evaluation pipelines can be exploited if not carefully designed. Future work could incorporate adversarial agents into the PoQ simulation, study robust aggregation rules that down-weight suspicious evaluators, and explore challenge-response mechanisms that combine PoQ with cryptographic verification techniques such as optimistic proofs or zero-knowledge arguments.

Finally, the study considers a fixed reward function with linear trade off between quality and cost. Alternative formulations, for example convex penalties on extreme costs or task-dependent weighting of quality metrics, may yield more nuanced incentive landscapes. Dynamic adjustment of reward coefficients based on network load or user preferences is another promising direction, especially for marketplaces where demand for different tasks fluctuates over time. Exploring these design choices in larger-scale simulations and, ultimately, in real blockchain-based deployments would provide deeper insight into the practicality of cost-aware PoQ as a foundation for decentralized LLM inference.

\section{Conclusion}

This paper presented a cost-aware extension of the Proof of Quality (PoQ) paradigm for decentralized large language model (LLM) inference. Instead of focusing solely on output correctness or raw model quality, the proposed design incorporates explicit measurements of computational cost into the reward function for both inference and evaluator nodes. A modular evaluation pipeline was introduced that combines ground truth metrics, lightweight learned evaluators, and GPT based judgments, enabling systematic analysis of quality signals and their alignment.

Experiments on extractive question answering and abstractive summarization with five instruction tuned LLMs and three evaluation models revealed several key insights. First, not all learned quality estimators are equally suitable for PoQ. Bi encoder models trained on semantic textual similarity, exemplified by STS--DistilRoBERTa, show substantially stronger correlation with both ground truth F1 and GPT scores than cross encoders tailored to retrieval or natural language inference. Second, quality–cost analysis demonstrated that larger models such as Llama-3.2-3B and Gemma-2-2B can simultaneously dominate smaller models in both accuracy and latency, underscoring the importance of empirical efficiency profiling rather than relying on parameter count as a proxy for cost. Third, Monte Carlo simulations indicated that the cost-aware reward function successfully shifts incentives toward high quality low cost inference models and toward evaluators that are both informative and efficient.

These findings suggest that cost-aware PoQ offers a promising foundation for economically sustainable decentralized LLM inference. By rewarding nodes according to quality-to-cost efficiency, the mechanism naturally promotes architectural choices and evaluator configurations that make effective use of available hardware while maintaining robust quality guarantees. Future extensions may incorporate heterogeneous devices, richer task distributions, and adversarial behaviors, moving closer to real-world blockchain-based deployments in which PoQ style mechanisms govern large-scale markets for LLM inference and evaluation services.

\bibliographystyle{plain}
\bibliography{references}

\end{document}